\documentclass[11pt]{article}

\usepackage[preprint]{acl}

\usepackage{times}
\usepackage{latexsym}

\usepackage[T1]{fontenc}
\usepackage[utf8]{inputenc}

\usepackage{microtype}
\usepackage{adjustbox}
\newcommand{\augrow}{AR & HS & PN & RS}

\usepackage{inconsolata}
\usepackage{booktabs}

\usepackage{graphicx}
\usepackage{xspace}
\usepackage{amsmath}
\usepackage{tikz}
\usepackage{seqsplit}
\usetikzlibrary{arrows.meta,positioning,fit,backgrounds,calc}
\usepackage{enumitem}

\newcommand{\framework}[0]{\textsc{alee}\xspace}

\title{\framework: Any-Language Evaluation of Embeddings via English-Centric Minimal Pairs}

\author{
    {\bf Andrianos Michail \hspace{2mm}}
    {\bf Stylianos Psychias \hspace{2mm}}
    {\bf Michelle Wastl \hspace{2mm}} \\
    {\bf Simon Clematide \hspace{2mm}} 
    {\bf Rico Sennrich \hspace{2mm}}
    {\bf Juri Opitz \hspace{2mm}} \\
    Department of Computational Linguistics \\
    University of Zurich \\
    \texttt{andrianos.michail@cl.uzh.ch}}

\begin{document}
\maketitle
\begin{abstract}

Text embeddings are standard for semantic similarity tasks, yet their evaluation remains an open challenge. Current benchmarks are static, cover only a limited set of languages, are often domain-specific, susceptible to overfitting, and poorly representative of low-resource languages. To address these limitations, we introduce \framework, a framework that extends \textit{Sentence Smith} \citep{li-etal-2025-sentence} to the cross-lingual and paragraph level. \framework uses Abstract Meaning Representations (AMR) to generate English minimal pairs with controlled, fine-grained semantic shifts, which are paired with translations in target languages. This approach enables targeted diagnostics for models in any language with English parallel data. We conduct a large-scale empirical study across a diverse set of embedding models and 275+ languages spanning three parallel datasets. On \framework, performance varies substantially across languages, text lengths, and linguistic phenomena, exposing persistent gaps in cross-lingual semantic representation that track language prevalence in training resources and subword tokenization. We release \textbf{\framework} at  \href{https://github.com/Andrian0s/any-lang-embed-eval}{https://github.com/Andrian0s/any-lang-embed-eval}.

\end{abstract}

\begin{figure}[t]
\centering
\begin{tikzpicture}[
node distance=1.4cm,
sentence/.style={
draw,
rounded corners,
fill=blue!5,
text width=0.95\columnwidth,
align=center,
inner sep=5pt
},
concept/.style={
draw,
rounded corners,
fill=gray!10,
minimum width=1.2cm,
minimum height=0.6cm,
font=\small
},
edge/.style={->, thick},
label/.style={font=\scriptsize},
amrbox/.style={draw, dashed, rounded corners, inner sep=8pt}
]

% --- Romansh sentence ---
\node[sentence] (rm) {
\small \textbf{Romansh – (Sursilvan variant)}\\
\textit{Il miedi crei che la terapia funcziuna.}
};

% --- English anchor ---
\node[sentence, below=0.3cm of rm] (en) {
\small  \textbf{Parallel English}\\
\textit{The doctor believes the treatment works.}
};

% --- AMR nodes (centered manually) ---
\node (center) at (en.south) {};
\node[concept] (believe) at ($(center)+(-0.6cm,-1.5cm)$) {believe-01}; 

\node[concept] (doctor) at ($(believe)+(-1.2cm,-1.2cm)$) {doctor}; 

\node[concept] (work) at ($(believe)+(1.2cm,-1.2cm)$) {work-01}; 

\node[concept] (treatment) at ($(work)+(0,-1.2cm)$) {treatment}; 

\node[concept, fill=red!10] (pol) at ($(believe)+(2.5cm,0cm)$) {--};

% --- AMR edges ---
\draw[edge] (believe) -- node[label,left,yshift=3pt] {:ARG0} (doctor.north);
\draw[edge] (believe) -- node[label,right,yshift=3pt] {:ARG1} (work.north);
\draw[edge] (work) -- node[label,right] {:ARG0} (treatment);

\draw[edge, red] (believe) -- node[label,above,xshift=-0.05cm] {:polarity} (pol);

% --- AMR box ---
\node[amrbox,
fit=(believe)(doctor)(work)(treatment)(pol),
label=above:{\textbf{AMR Manipulation}}
] (amr) {};

% --- Edited sentence ---
\node[sentence, below=0.9cm of amr] (neg) {
\small \textbf{Edited English}\\
\textit{The doctor doesn't believe the treatment works.}
};

% --- Pipeline arrows (perfectly vertical) ---
\draw[->, dashed] (en.south) -- node[label, right]{\textbf{PARSE}} (en.south |- amr.north);
\draw[->, dashed] (amr.south) -- node[label, right]{\textbf{GENERATE}} (amr.south |- neg.north);

\end{tikzpicture}

\caption{Controlled cross-lingual polarity flip in \framework.}

\label{fig:polarity_romansh}

\end{figure}
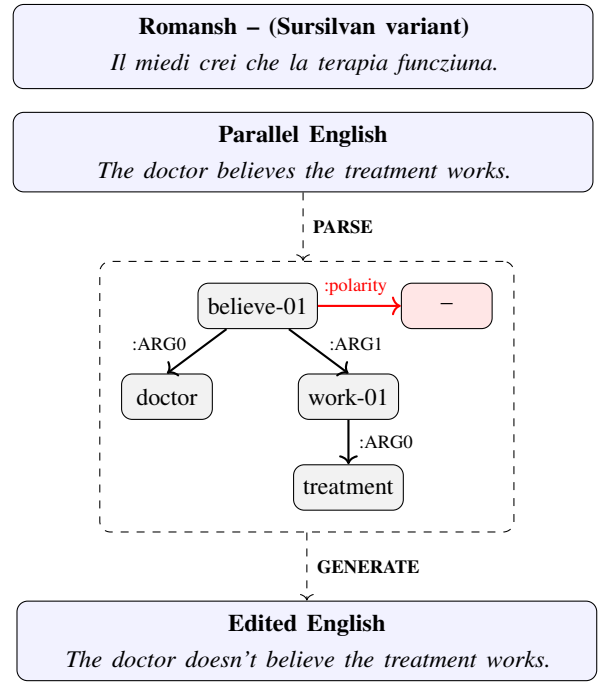

\section{Introduction}

Semantic text embeddings are central to modern information retrieval, clustering, and cross-lingual alignment \citep{reimers-gurevych-2019-sentence,gao-etal-2021-simcse}. However, their evaluation is largely based on static benchmarks with coarse-grained similarity judgments \citep{muennighoff-etal-2023-mteb}. These datasets have three major limitations: they are  biased toward high-resource languages, they are vulnerable to data leakage and overfitting because they are fixed, and they are too coarse-grained to distinguish semantic equivalence from lexical overlap. 

To address these limitations, we propose a dynamic minimal-pair evaluation setting leveraging Abstract Meaning Representation \citep[AMR,][]{banarescu2013abstract}. AMR is a formal semantic representation of a text's meaning, explicating semantic phenomena such as \textit{entities} and their \textit{roles}, \textit{negation}, \textit{cause}, and more---making it suitable for controlled representation and generation tasks \cite{wein-opitz-2024-survey,sadeddine-etal-2024-survey}. Instead of relying on fixed sentence pairs as in previous embedding evaluations, we use AMR to generate challenging sentence examples that are lexically and syntactically close to a source sentence but differ by one controlled semantic operation.  

Figure~\ref{fig:polarity_romansh} illustrates the minimal pair generation with a sentence in Sursilvan, a Romansh variety spoken in the Swiss canton of Graubünden~\citep{Moseley2010-sr}. The English parallel sentence, ``The doctor believes the treatment works'' is parsed into an AMR graph, and a \texttt{:polarity -} edge is added to the predicate \textit{believe-01}. This yields the sentence ``The doctor doesn't believe the treatment works,'' while the Sursilvan sentence remains unchanged: ``Il miedi crei che la terapia funcziuna''. Henceforth, we denote such a generation also by \textit{foil} or \textit{confounder}, as this indicates its functional purpose within our embedding evaluation framework. 

This construction allows us to test whether a model assigns higher similarity to the original English sentence than to its minimally perturbed foil, in relation to the same target-language sentence.

Because the English sentences differ only in one controlled semantic feature, a model's preference for the original over the foil provides a targeted diagnostic of its cross-lingual embedding behavior. This way, we can test whether target-language representations are aligned closely enough with English to preserve distinctions such as polarity, argument structure, or lexical-semantic contrast.

In this paper, we introduce \framework (Any-Language Evaluation of Embeddings), a framework that combines English-side semantic control with broad language coverage. By generating English minimal pairs and pairing them with target-language data, we create a diagnostic testbed for any language with English parallel data. We show the value of \framework by evaluating a diverse set of embedding models across 275+ languages, including very low-resource varieties, and report consistent patterns.

\noindent\textbf{Our contributions and findings:}
\begin{enumerate}[nosep,leftmargin=*]
\item We introduce \framework, a dynamic cross-lingual benchmark construction framework based on English AMR-derived minimal pairs and target-language parallel data.
\item Even strong embedding models fail to resolve all minimal pairs, with structural and lexical shifts such as role reversals, antonymy, and abstraction-level changes proving harder than polarity negation; notably, larger decoder-based models do not outperform smaller encoders.
Furthermore, longer, multi-sentence texts are substantially harder than shorter ones.%, even when only one sentence is perturbed.
\item \framework performance correlates with language prevalence in training corpora:
%\item We find evidence that 
both pretraining and fine-tuning language distributions correlate with per-language performance, with fine-tuning more strongly associated.
\item Greater subtoken fragmentation correlates strongly with lower downstream performance.
\item In a Romansh case study, performance declines across varieties in roughly the order of speaker population, mirroring written prevalence and tokenizer coverage down to the dialect level.
\end{enumerate}

\begin{table*}[ht]
\centering
\small % Slightly smaller to fit the descriptive column
\begin{tabular}{@{}llll@{}}
\toprule
\textbf{Transformation} & \textbf{Linguistic Target} & \textbf{Relation Induced} & \textbf{Description} \\ \midrule
\textbf{PN} & Truth-functional Logic & Contradictory / Contrary & \begin{tabular}[c]{@{}l@{}}Negates via \textit{:polarity -}. Distinguishes \\ Syntactic and morphological opposition.\end{tabular} \\ \addlinespace
\textbf{RS} & Argument Structure & Reversal of Thematic Roles & \begin{tabular}[c]{@{}l@{}} Swaps :ARG0 (Agent) and :ARG1 (Patient). \\Tests sensitivity to semantic structure.\end{tabular} \\ \addlinespace
\textbf{AR} & Lexical Semantics & Polar Opposition & \begin{tabular}[c]{@{}l@{}}Replaces node with WordNet antonym. \\ Targets conceptual opposites.\end{tabular} \\ \addlinespace
\textbf{HS} & Taxonomic Hierarchy & Hyponymy / Entailment & \begin{tabular}[c]{@{}l@{}}Replaces concept with superordinate. \\ Breaks bidirectional entailment.\end{tabular} \\ \bottomrule
\end{tabular}
\caption{Mapping AMR Transformations to Linguistic Targets and Semantic Relations.}
\label{tab:amr-transformations}
\end{table*}

\section{Related Work}

\paragraph{Cross-Lingual Embedding Evaluation.}
Common evaluation tasks are cross-lingual information retrieval \citep[CLIR;][]{lawrie2025neuclirbenchmodernevaluationcollection}, semantic text similarity \citep[X-STS;][]{cer2017semeval}, and bitext mining \citep{zweigenbaum2017overview}. These evaluations provide a useful overall view of model quality, but they offer limited insight into which semantic distinctions a model captures across languages. Strong results may arise from broad lexical or topical overlap, underweighting sensitivity to finer meaning contrasts \citep{fodor-etal-2025-compositionality}. Further, the staticness of datasets risks overfitting effects, and the concentration on higher-resource languages makes them less-suitable for fine-grained diagnosis in low-resource settings, and unsuitable for languages not covered, such as the Romansh varieties that we cover in our work here.

Embedding evaluation also ties closely to embedding interpretability  \citep{opitz-etal-2025-interpretable,opitz-etal-2026-similar}. A recurring challenge is to move beyond holistic similarity scores and identify which semantic properties account for textual differences. Our work aligns with this goal by using controlled semantic edits for targeted probes of cross-lingual spaces.% in a targeted manner.% test semantic dimensions %, such as  negation or role reversals,
%in multilingual embedding spaces.% \citep{zhu-etal-2018-exploring,zhu-de-melo-2020-sentence}.

\paragraph{Minimal Pairs} aim to isolate specific linguistic contrasts, but their datasets require costly manual effort and are static \citep[e.g., BLiMP and Multi-BLiMP;][]{warstadt-etal-2020-blimp-benchmark,10.1162/TACL.a.600}. By contrast,
\textit{Sentence Smith} \citep{li-etal-2025-sentence} generates English minimal pairs through rule-based edits to AMR graphs as an intermediate representation---in a dynamic fashion that allows ad-hoc generation of novel pairs. \framework extends this idea cross-lingually by applying English-side perturbations to parallel corpora. Unlike other work relying on black-box LLM verification or costly human data creation \citep{nastase-etal-2024-exploring, michail-etal-2025-examining}, we use explicit semantic representation and can scale to any language with English parallel data.

\section{\framework}

We use \framework to denote both the dataset construction pipeline and  the resulting evaluation datasets. 

\subsection{Controlled Foil Generation via AMR}
To generate semantically controlled minimal pairs, we build on \textit{Sentence Smith}  \citep{li-etal-2025-sentence}. Its pipeline consists of three stages: (1) parsing an English source sentence into an AMR graph, (2) applying a rule-based semantic edit to the graph, and (3) generating text from the modified graph.

\framework extends this process to a cross-lingual setting by applying the edit to the English side of a parallel corpus. This yields triplets of the form $(\text{en}_{\text{orig}}, \text{en}_{\text{foil}}, \text{target}_{\text{orig}})$. We then evaluate whether a model assigns higher similarity to the original cross-lingual pair $(\text{en}_{\text{orig}}, \text{target}_{\text{orig}})$ than to the foil pair $(\text{en}_{\text{foil}}, \text{target}_{\text{orig}})$.

\subsection{Semantic Manipulation Types}
Following \citet{li-etal-2025-sentence}, we apply four AMR-based perturbation types to English pivot texts. Each targets a different type of semantic distinction, as summarized in Table~\ref{tab:amr-transformations}.

\begin{itemize}[itemsep=0em,leftmargin=15pt]
    \item \textbf{Polarity Negation (PN):} This manipulation reverses the truth value of a predicate by adding the \texttt{:polarity -} attribute. It covers cases of \textbf{syntactic negation} (e.g., \textit{approve} $\rightarrow$ \textit{not approve}), which functions as a logical \textbf{contradictory} (absence of a property), and in some cases \textbf{morphological negation} (e.g., \textit{just} $\rightarrow$ \textit{unjust}), although the latter is not always strictly equivalent to simple negation. 
    
    \item \textbf{Role Swap (RS):} This manipulation perturbs \textbf{predicate-argument structure} by exchanging  thematic roles such as Agent and Patient. By swapping the \texttt{:ARG0} and \texttt{:ARG1} nodes, it changes \textit{who did what to whom} while preserving most lexical material, thereby testing sensitivity to structure rather than surface overlap.

    \item \textbf{Antonym Replacement (AR):} This manipulation substitutes a concept with a WordNet-derived \textbf{antonym}, e.g., \textit{good} $\rightarrow$ \textit{bad}. Unlike polarity negation, it relies on lexical substitution rather than a negation marker, thereby probing sensitivity to lexical-semantic opposition. 

    \item \textbf{Hypernym Substitution (HS):} This manipulation replaces a concept with its taxonomic \textbf{superordinate} (e.g., \textit{penguin} $\rightarrow$ \textit{bird}). Because the resulting sentence is semantically less specific, this perturbation tests whether embeddings preserve distinctions involving lexical entailment relations and level of abstraction.
\end{itemize}

\subsection{Validation of Generated Foils}
\label{sec:validation}

Not every graph manipulation might yield a correct foil: in some cases, the generated sentence could remain too close in meaning to the source or the semantic change could be neutralized during generation, e.g., by the AMR generator failing to express our change. We therefore apply a bidirectional entailment filter using the robustified NLI model of \citet{steen-etal-2023-little}, adding an additional verification step that is efficient and fully automated.

For each source-foil pair, we run the NLI model in both directions: source $\rightarrow$ foil and foil $\rightarrow$ source. To reduce false positives, we reject a candidate if (i) both directions are predicted as entailment, or (ii) P(entail) in either direction exceeds 0.8. The latter may be seen as standing in some conflict with the desired HS-relation, but practical experiments with the specific NLI module indicated (i) and (ii) as a suitable setup for minimizing false positives. %This criterion is applied uniformly across all manipulation types and serves only as a filtering heuristic: it does not verify the semantic effect targeted by each manipulation (Table~\ref{tab:amr-transformations}).

\subsection{Iterative Paragraph Manipulation}
\label{sec:iterative}
AMR-based generation works best at the sentence level. To handle the paragraph-length texts in datasets such as WMT24++, we propose an iterative first-success procedure:
\begin{enumerate}[itemsep=0em,leftmargin=15pt]
    \item \textbf{Decomposition:} The paragraph is segmented into individual sentences using NLTK's sent\_tokenize(v3.9).
    \item \textbf{Surgical Edit:} We attempt an AMR manipulation on the $i$-th sentence (start with $i$=0). 
    \item \textbf{Sentence Validation:} The modified sentence is checked through an NLI check against the original sentence as described in Section \ref{sec:validation}.
    \item \textbf{Construction \& Validation:} The edited sentence is integrated in the paragraph. An NLI check is also run against the original paragraph. 
    \item \textbf{Recursion:} If the NLI model fails to confirm a semantic shift at either stage (i.e., the paragraphs remain bi-directionally entailed), we revert the change, increment $i$, return to step 2.
\end{enumerate}

\paragraph{Discussion.} These four perturbation types allow us to evaluate models along distinct semantic dimensions for any language paired with English parallel data. In the cross-lingual setting, the model must determine which of two closely related English sentences better matches a target-language sentence. In the next section, we describe the parallel datasets used to instantiate this framework.

\subsection{Multilingual Corpora}
We instantiate \framework on three parallel datasets to provide broad linguistic and structural coverage. 

\paragraph{\textit{\textbf{F}LORES-200}} is a widely used Machine Translation test set with professionally translated sentences in \textbf{200 languages} \citep{nllb2022arxiv}. We refer to this dataset instance as \textbf{\framework-F200}.

\paragraph{\textit{W\textbf{MT}24++}} is an expanded parallel corpus containing 998 texts in 55 languages \citep{deutsch-etal-2025-wmt24}; a later extension adds six Romansh varieties, yielding \textbf{61 languages} in total \citep{vamvas-etal-2025-expanding}. Because WMT24++ includes both sentence-level and paragraph-level material, it also utilizes the iterative editing pipeline described above. We refer to this dataset instance as \textbf{\framework-MT61}.

\paragraph{\textit{\textbf{B}OU\textbf{Q}uET}} is a multi-way parallel, multi-register benchmark whose source texts were handcrafted by linguists in 8 diverse languages and translated into English and 266 additional language varieties, yielding \textbf{275 languages} in total \citep{andrews-etal-2025-bouquet, omnilingual2026}. Because BOUQuET, like WMT24++, provides both sentence-level and paragraph-level material, it also utilizes the iterative editing pipeline. We refer to this dataset instance as \textbf{\framework-BQ275}.

\begin{table}[t]
\centering
\small
\begin{tabular}{lccc}
\toprule
 & \textbf{\framework} & \textbf{\framework} & \textbf{\framework} \\
 & \textbf{F200} & \textbf{MT61} & \textbf{BQ275} \\
\midrule
English sources & 1,012 & 818 & 1,052 \\
\midrule
Polarity negation & 558 & 668 & 669 \\
Role swap & 334 & 377 & 333 \\
Antonym replacement & 484 & 609 & 597 \\
Hypernym substitution & 216 & 354 & 341 \\
\bottomrule
\end{tabular}
\caption{Count of sources and NLI-validated foils.}
\label{tab:dataset_stats}
\vspace{-0.5\baselineskip}
\end{table}

\paragraph{\textbf{\framework-F200} Statistics.} We apply all four transformations to 1,012 English source sentences from FLORES-200. Table~\ref{tab:dataset_stats} reports the number of validated foils per transformation. The median source length is 24 tokens (IQR: 19–29), and the median foil length is 20 tokens (IQR: 15–26).

\paragraph{\textbf{\framework-MT61} Statistics.} We exclude 180 (out of 998) flagged as low quality by the corpus providers, leaving 818 texts for foil generation. Success rates are higher than in \textbf{\framework-F200} (Table~\ref{tab:dataset_stats}), likely because paragraph-level texts offer multiple opportunities to obtain at least one valid edit. The median source length is 38 tokens (IQR: 17–73), and the median foil length is 44 tokens (IQR: 19–76).

\paragraph{\textbf{\framework-BQ275} Statistics.} We apply all four transformations to 854 sentences and 198 paragraphs from BOUQuET; Table~\ref{tab:dataset_stats} reports the amount of yielded foils. Median source length is 15 tokens (IQR: 10–16); median foil length 14 (IQR: 9–14).

\subsection{Quality Assessment}

We manually inspect 200 English source--foil pairs sampled from each dataset. In \textbf{\framework-F200}, all 200 pairs are judged to be valid minimal pairs, and 98\% are judged grammatical. 
Most grammaticality errors arise from local syntactic artifacts that resemble poorly integrated template insertions. 
For the 200 sampled pairs from \textbf{\framework-MT61}, we obtain similarly high quality. We judge 99\% of the minimal pairs to be valid and 96\% to be grammatical. The remaining errors are mainly due to formatting artifacts, such as HTML tags, or punctuation errors. For \textbf{\framework-BQ275} all sampled minimal pairs are considered valid, while 98\% are grammatical. Overall, this manual assessment is consistent with the high data quality reported by \citet{li-etal-2025-sentence}.

Note that role-swapped constructions occasionally do not adhere to selectional preferences, i.e., the semantic constraints that predicates impose on their arguments~\citep{Wilks1975-qg,dowty1991thematic}. E.g.: \\[4pt]
\textbf{orig:} \textit{The effect the team was looking for would be caused by tidal forces between the galaxy's dark matter and the Milky Way's dark matter.}\\
\textbf{foil:} \textit{The effects that the tide looks at are caused by the team forcing the dark matter of the galaxy to darken.} 
\\[4pt]
Yet, we emphasize that even in those cases, the main aim of the operation has been achieved. The \textbf{foil} in this case may appear nonsensical or odd, but it is a valid test case. %Hence, these serve as a valid and rigorous test of whether models rely more on selectional heuristics than on semantic content.

\section{General Setup}

\paragraph{Evaluation Metric.}

We follow \citet{li-etal-2025-sentence} and use Triplet Accuracy (\textit{TACC}), defined as the proportion of triplets in which a model assigns higher similarity to the original cross-lingual pair than to the foil pair. When aggregating over perturbation types, we report \textbf{Macro-TACC}, which averages \textbf{TACC} across the four augmentations. Higher values are better; random performance is 0.5.

\begin{figure}[h!]
\centering
\includegraphics[width=0.92\linewidth]{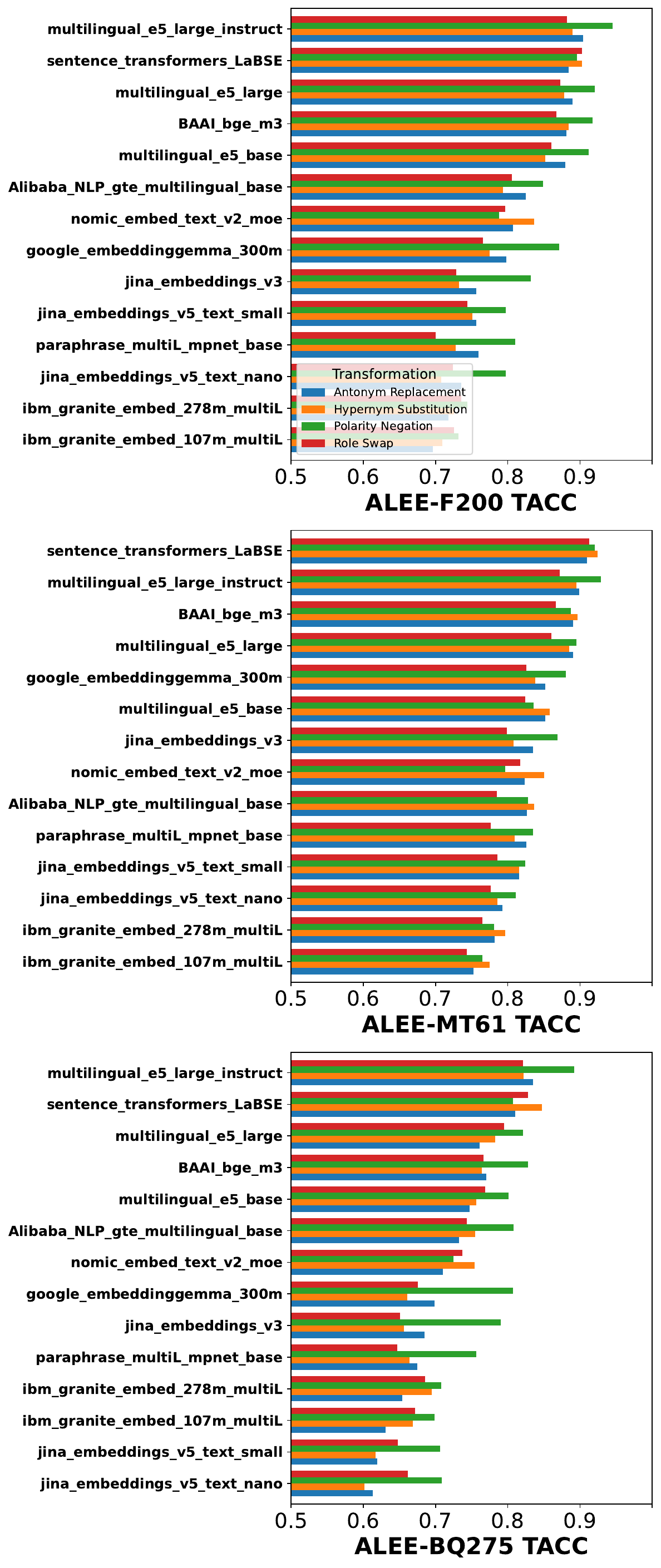}
\caption{Mean TACC per model broken down by semantic edit type, on \textbf{\framework-F200} (top), \textbf{\framework-MT61} (middle) and \textbf{\framework-BQ275 (bottom)}. Models are sorted by averages across minimal pair types.}
\label{fig:aug_tacc_barh}
\end{figure}

\begin{figure*}[t]
\centering
\includegraphics[width=\textwidth]{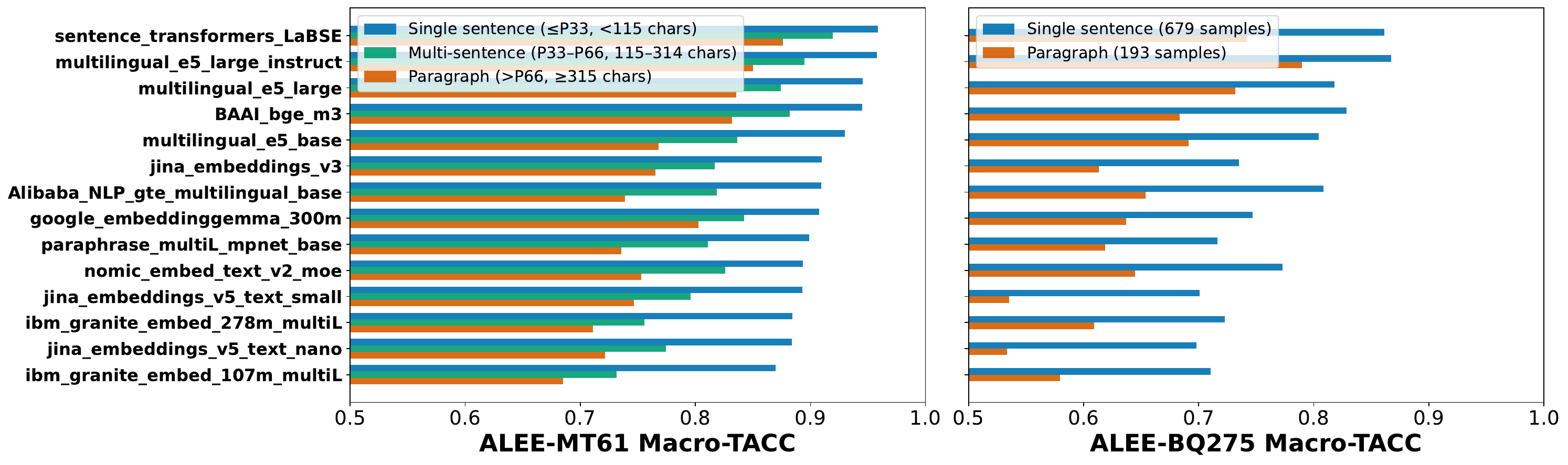}
\caption{\textbf{\framework-MT61, \framework-BQ275} Macro-TACC across different text lengths.}
\label{fig:text-len-distribution-performance}
\end{figure*}

\paragraph{Examined Embedding Models.}

We evaluate a set of widely used embedding models with fewer than 600M parameters: the bitext-mining model \texttt{\seqsplit{sentence\_transformers\_LaBSE}} \citep{feng-etal-2022-languagelabse}, the paraphrase-aligned \texttt{\seqsplit{paraphrase\_multiL\_mpnet\_base}} \citep{reimers-gurevych-2019-sentence, reimers-gurevych-2020-making}, Google's \texttt{\seqsplit{google\_embeddinggemma\_300m}} \citep{vera2025embeddinggemmapowerfullightweighttext}, \texttt{\seqsplit{nomic\_embed\_text\_v2\_moe}} \citep{nussbaum2025trainingsparsemixtureexperts}, \texttt{\seqsplit{Alibaba\_NLP\_gte\_multilingual\_base}} \citep{zhang-etal-2024-mgte}, \texttt{\seqsplit{BAAI\_bge\_m3}} in dense mode \citep{chen-etal-2024-m3}, the \texttt{\seqsplit{jina\_embeddings\_v3}} and \texttt{\seqsplit{jina\_embeddings\_v5\_text\_small/nano}} models \citep{gunther_jina_v3, akram2026jinaembeddingsv5texttasktargetedembeddingdistillation}, \texttt{\seqsplit{ibm\_granite\_embed\_107/278m\_multiL}} \citep{awasthy2025graniteembeddingmodels}, and the Multilingual E5 family---\texttt{\seqsplit{multilingual\_e5\_base/large/instruct}} \citep{wang2024multilinguale5textembeddings}. Detailed configurations are provided in Appendix Table~\ref{tab:model-configs}.
\section{Research Findings}

\subsection{Fine-Grained Linguistic Evaluation}

\paragraph{Research Question.} We first ask how embedding models perform on \framework overall and how their performance varies across the four semantic perturbation types.

\paragraph{Setup.} For each model and each transformation, we compute TACC averaged over languages, separately for each \framework dataset.

\paragraph{Findings.} Figure~\ref{fig:aug_tacc_barh} summarizes the results. We observe three main patterns: 
(1) Multilingual E5-instruct and LaBSE rank at or near the top on all three datasets. Multilingual E5-instruct leads on \textbf{\framework-F200} and \textbf{\framework-BQ275}, while LaBSE---a model trained for bitext mining with hard negatives---edges ahead on \textbf{\framework-MT61}. At the same time, no model reaches perfect accuracy, indicating that all models fail on a non-trivial subset of minimal pairs.
(2) Across all three datasets, polarity negation is consistently the easiest perturbation, whereas role swap, antonym replacement, and hypernym substitution are closely clustered and show no stable ordering across datasets. This suggests that models detect explicit polarity reversals more reliably than the other contrasts, which they handle at broadly similar levels.
(3) In aggregate, performance is highest on \textbf{\framework-MT61}, followed by \textbf{\framework-F200} and \textbf{\framework-BQ275} (Table~\ref{tab:qwen_appendix}). We attribute this ordering to language coverage rather than text difficulty: \textbf{\framework-F200} and \textbf{\framework-BQ275} include more low-resource languages, which lowers their averaged scores. Restricting the comparison to languages shared across all three datasets removes this effect.
Full per-language results are in Appendix~\ref{app_per_language}, where a language can be located by searching (\texttt{Ctrl+F}) for its three-letter code followed by an underscore (e.g., ``\texttt{deu\_}'' for German).

\subsection{Effects of Text Length}

\paragraph{Research Question.} A key extension of \textit{Sentence Smith} in \framework is the ability to manipulate longer, multi-sentence texts through the iterative sentence processing pipeline described in Section~\ref{sec:iterative}. We therefore examine how model performance varies with text length, averaged across all languages.

\paragraph{Setup.} In \textbf{\framework-MT61}, we divide texts into three bins based on character-length percentiles (below P33, between P33 and P66, and above P66) and compute the average Macro-TACC for each model in each bin. In \textbf{\framework-BQ275}, we use the provided sentence or paragraph level flag and compute the average Macro-TACC for each model in each bin.

\begin{figure*}[t]
\centering
\includegraphics[width=0.98\textwidth, trim=20 0 20 0, clip]{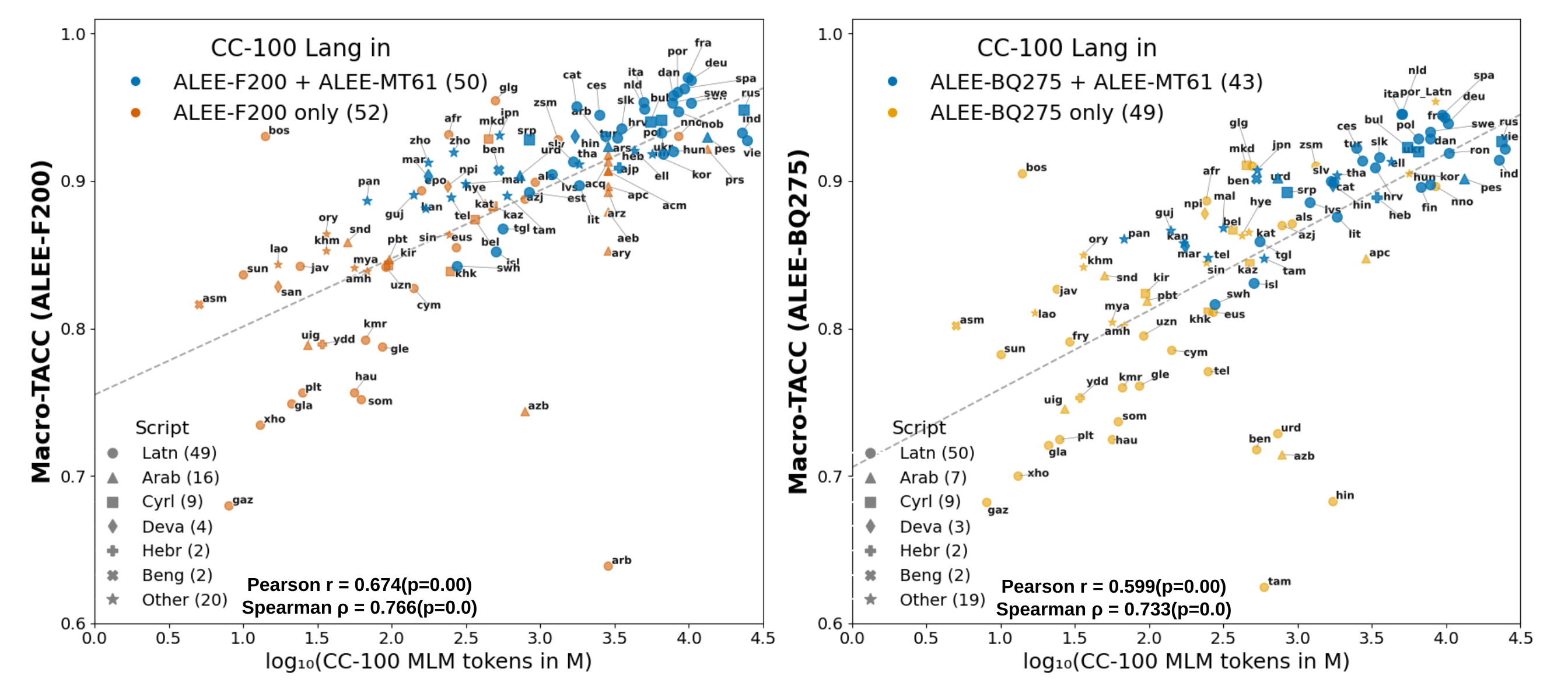}
\caption{Per-language Macro-TACC vs.\ CC-100 pretraining data size (log-scaled), averaged across all models, for \textbf{\framework-F200} (left) and \textbf{\framework-BQ275} (right). Each point is a language, colored by whether it also appears in \textbf{\framework-MT61}. Marker shapes indicate writing script.}
\label{fig:flores_cc100_scatter_scripts}
\end{figure*}

\begin{figure}
\centering
\includegraphics[width=0.99\linewidth, trim=26 0 26 0, clip]{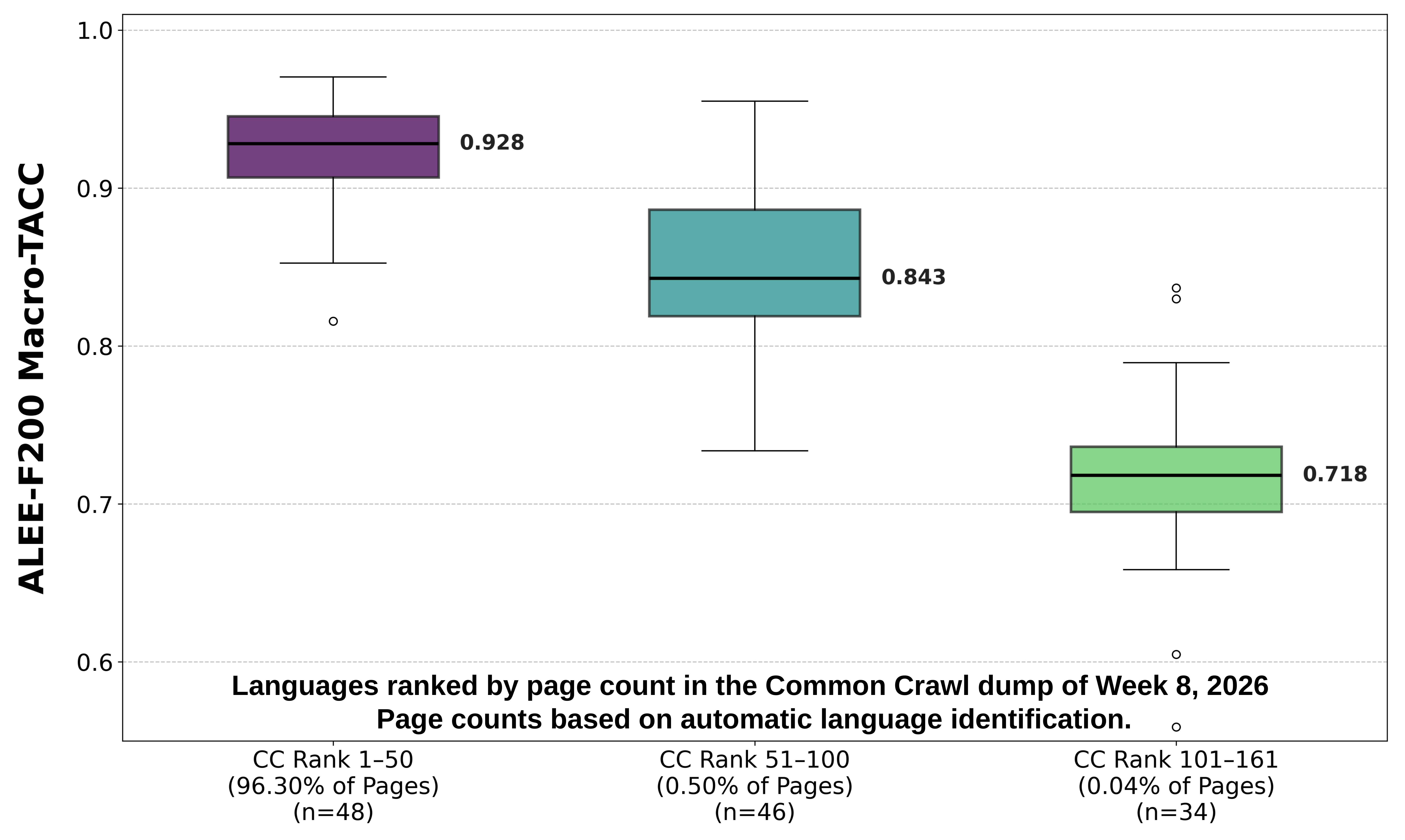}
\caption{\textbf{\framework-F200} Macro-TACC distribution by Common Crawl language prevalence. \textbf{\framework-MT61} and \textbf{\framework-BQ275} show similar patterns (Figs.~\ref{fig:resource_group_boxplot_mt61} and~\ref{fig:resource_group_boxplot_bq275}).}
\label{fig:resource_group_boxplot}
\vspace{-0.5\baselineskip}
\end{figure}

\paragraph{Findings.} Figure~\ref{fig:text-len-distribution-performance} shows a consistent decline in performance as text length increases, even though each foil modifies only a single sentence. In \textbf{\framework-MT61}, the average drop across models is 9.4\% from single- to multi-sentence texts and 15.4\% from single-sentence to paragraph. Within \textbf{\framework-BQ275}, performance drops similarly by 16.0\% from sentence to paragraph level. This suggests that semantic discrimination becomes more difficult as the amount of surrounding context grows.%, perhaps aggravated by certain long-text biases \citep{schuhmacher2026informationrepresentationfairnesslongdocument}.

\begin{figure*}[t]
\centering
\includegraphics[width=0.923\textwidth, trim=5.8 0 5.8 0, clip]{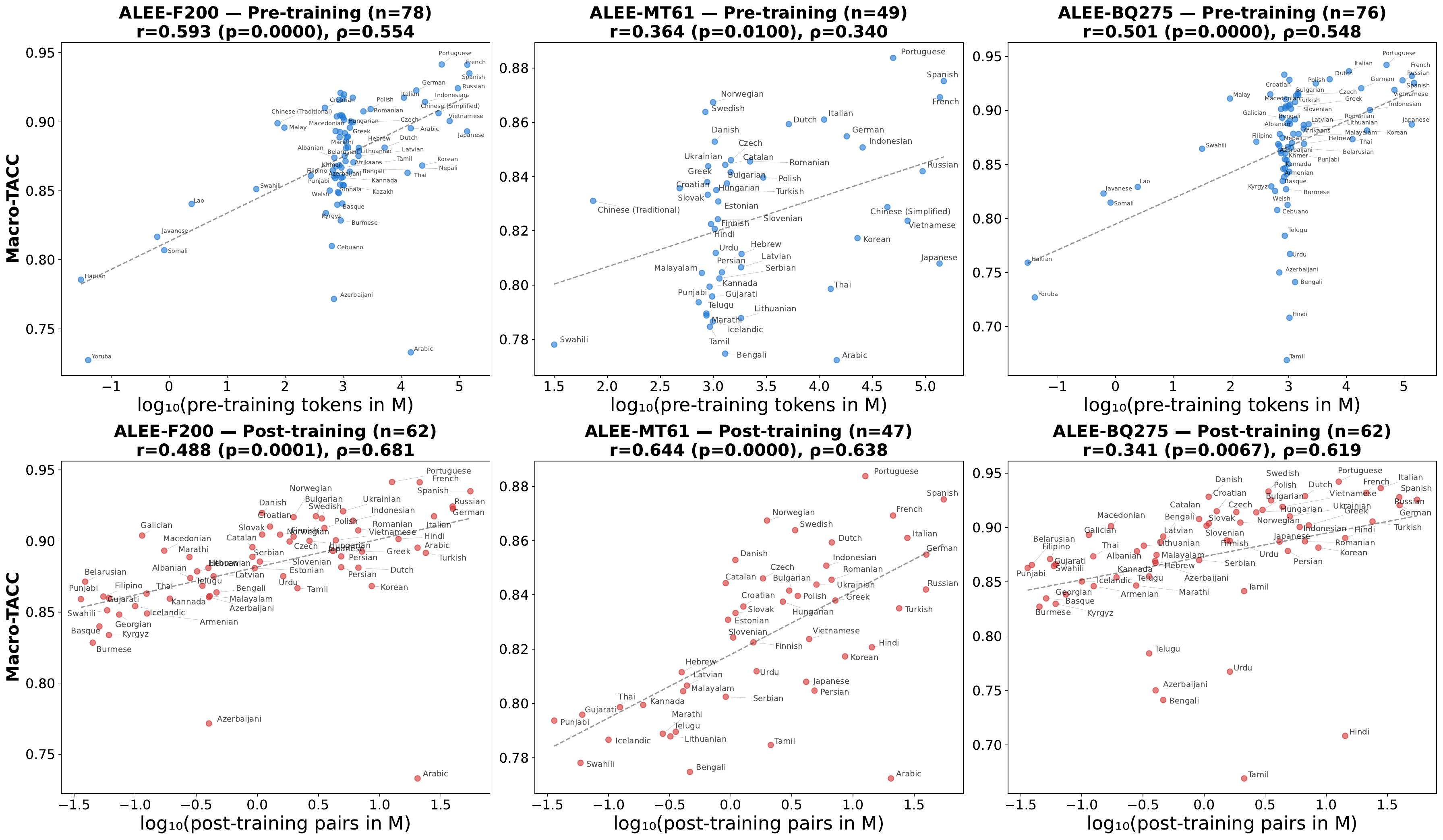}
\caption{multilingual-gte-base. Per-language TACC vs.\ training data size for pretraining tokens (top) and post-training pairs (bottom). $r/\rho$ Pearson's/Spearman's coefficient.}
\label{fig:gte_combined_2x3}
\end{figure*}

\subsection{Training Language Distribution Effects}

\paragraph{Research Question.} We now investigate which language-level factors are associated with performance differences across languages. To this aim, we first  investigate the language distribution of a widely used pretraining corpus, and finally aim to disentangle the contributions of pretraining and fine-tuning data for an exemplar model.

\subsubsection{Pretraining Data Distribution}

\paragraph{Setup.} To move beyond coarse resource membership, we examine the relationship between pretraining data size and downstream performance. Because 5 of the 14 encoder models use XLM-RoBERTa \citep{conneau-etal-2020-unsupervised} as a backbone and 3 others reuse its tokenizer, we use CC-100 language proportions as an approximate proxy for pretraining exposure. For each language reported in CC-100, we compare log-scaled data size with Macro-TACC averaged across models, separately for \textbf{\framework-F200} and \textbf{\framework-BQ275}.

\paragraph{Findings.} Figure~\ref{fig:flores_cc100_scatter_scripts} shows a strong positive correlation between CC-100 data size and performance on both datasets (Spearman $\rho$=0.766 on \textbf{\framework-F200}; $\rho$=0.733 on \textbf{\framework-BQ275}). This trend is consistent with the broader pattern: languages with greater representation in pretraining resources tend to achieve better embedding performance.

Note that this finding is tied to the models with the same backbone. To further corroborate this finding from a broader perspective, we assess performance of all embedding models on different subsets of languages, adopting the assumption that a low-resource language is a language that occurs less frequently on the internet. The result in Figure \ref{fig:resource_group_boxplot} clearly corroborates the strong relationship between embedding performance and language representation in online texts.

\subsubsection{MLM vs.\ Contrastive Training}

\paragraph{Setup.} To distinguish the contributions of pretraining and fine-tuning, we examine GTE Multilingual Base, which openly reports its MLM pretraining and contrastive fine-tuning data distributions per language. For each language, we plot its Macro-TACC against the log-scaled data volumes, separately for each \framework dataset.

\paragraph{Findings.} Figure~\ref{fig:gte_combined_2x3} shows that both training stages correlate positively with performance across all three datasets, with contrastive fine-tuning data more strongly associated with performance than MLM pretraining data. This gap varies by dataset: it is largest on \textbf{\framework-MT61} ($\rho$=0.638 vs $\rho$=0.340), and smaller on \textbf{\framework-F200} ($\rho$=0.681 vs $\rho$=0.554) and \textbf{\framework-BQ275} ($\rho$=0.619 vs $\rho$=0.548). This may be because \textbf{\framework-MT61} contains longer texts and thus provides a stronger out-of-distribution test than \textbf{\framework-F200} and \textbf{\framework-BQ275}.

Across all three correlation analyses, performance is consistently associated with language representation in the training pipeline, including broad corpus coverage and training volumes.

\subsection{Subword Tokenization Effects}

\paragraph{Setup.} As a corpus-independent proxy for language coverage, we compute the median number of XLM-R subtokens per language and correlate it with the average TACC over all models. A higher number of subtokens may reflect not only limited vocabulary coverage, but also morphological complexity or script-related properties.

\begin{figure}[t]
  \centering
  \includegraphics[width=0.99\linewidth, trim={0 1.36cm 0.95cm 0}, clip]{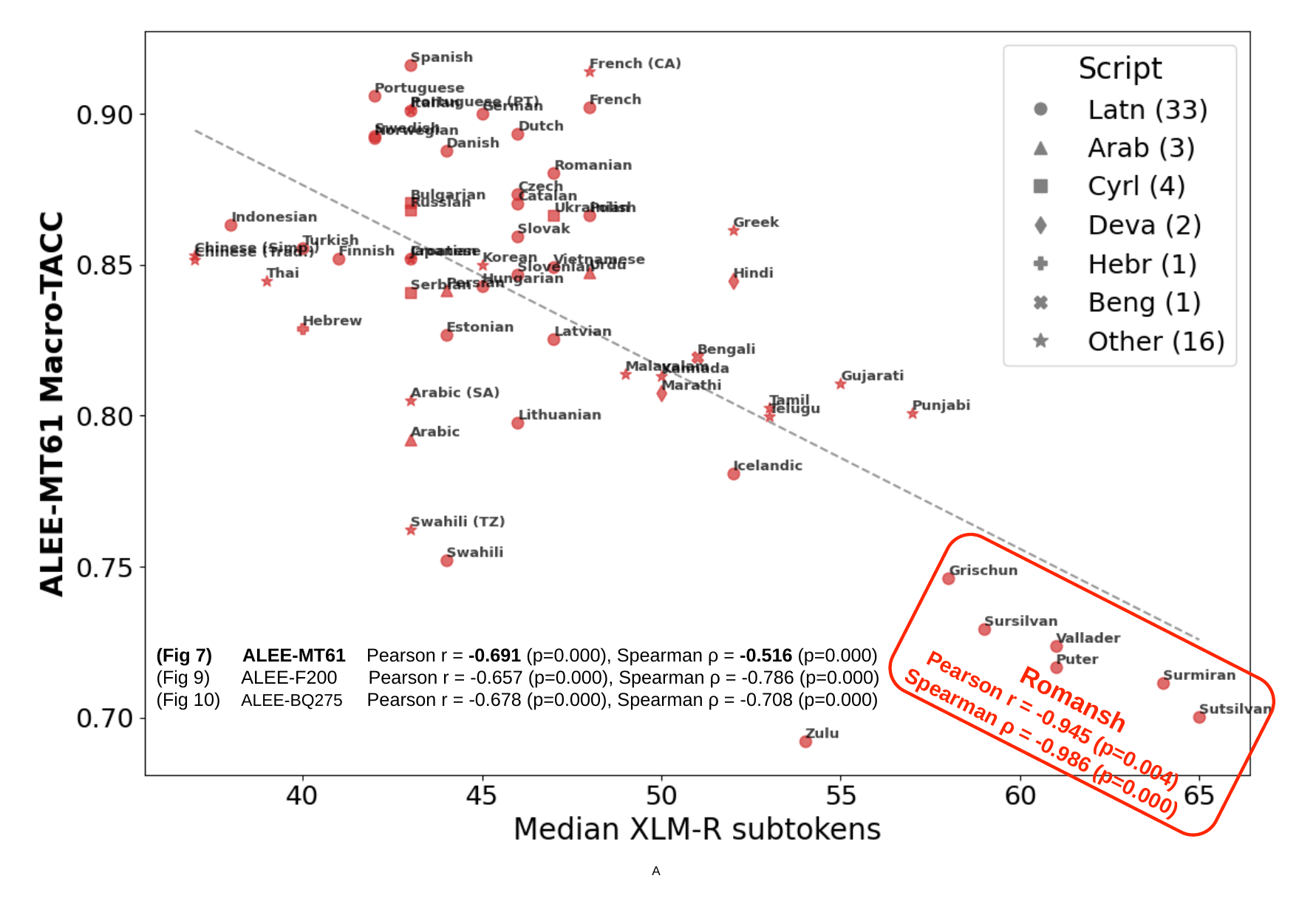}
  \caption{Per-language Macro-TACC vs.\ median XLM-R subtoken count on \textbf{\framework-MT61}, averaged across all models. Marker shapes indicate writing script. Figures~\ref{fig:subtokens_f200} \& \ref{fig:subtokens_bq275} in the Appendix show \textbf{\framework-F200/BQ275} results.}
  \label{fig:performance_to_subtokens}
\end{figure}

\paragraph{Findings.} Figure~\ref{fig:performance_to_subtokens} shows the results for \textbf{\framework-MT61}; Results \textbf{\framework-F200} and \textbf{\framework-BQ275} are reported in the Appendix Figures~\ref{fig:subtokens_f200} and \ref{fig:subtokens_bq275}.

Languages whose texts are split into more subtokens tend to show lower performance ($\rho$=$-$0.516 on \textbf{\framework-MT61}; $\rho$=$-$0.786 on \textbf{\framework-F200}, $\rho$=$-$0.708 on \textbf{\framework-BQ275}). The Romansh idioms illustrate this pattern perfectly: they require the largest number of subtokens and show the lowest performance, with an almost perfectly negative correlation across the six idioms ($\rho$=$-$0.986).

Overall, greater subword fragmentation is associated with lower embedding performance, whether it arises from limited vocabulary coverage or language-specific structural properties.

\begin{table}[t]
\centering
\small
\setlength{\tabcolsep}{4pt}
\resizebox{\linewidth}{!}{%
\begin{tabular}{lccccc}
\toprule
 & \textbf{Rank} & \textbf{\framework} & \textbf{\framework} & \textbf{\framework} & \textbf{Avg} \\
 & & \textbf{F200} & \textbf{MT61} & \textbf{BQ275} & \\
\midrule
LaBSE & 2 & 0.896 & 0.917 & 0.823 & 0.879 \\
\textbf{Qwen3-Emb-8B} & 6 & 0.842 & 0.839 & 0.723 & 0.801 \\
gte-multilingual-base & 7 & 0.818 & 0.819 & 0.760 & 0.799 \\
\textbf{Qwen3-Emb-4B} & 11 & 0.777 & 0.799 & 0.680 & 0.752 \\
granite\_embed107\_multiL & 16 & 0.716 & 0.759 & 0.667 & 0.714 \\
\textbf{Qwen3-Emb-0.6B} & 17 & 0.696 & 0.735 & 0.644 & 0.691 \\
\bottomrule
\end{tabular}%
}
\caption{Overall and per-dataset Macro-TACC for the Qwen decoder-based models and selected comparison models. Rank is over all evaluated models on the three-dataset average.}
\label{tab:qwen_main}
\vspace{-0.5\baselineskip}
\end{table}

%\begin{figure*}[t]
%\centering
%\begin{minipage}{0.49\textwidth}
%  \centering
%  \includegraphics[width=0.98\linewidth, trim={0 0.5cm 0.5cm 0}, clip]{subtoken_mt61.pdf}
%  \caption{Per-language Macro-TACC vs.\ median XLM-R subtoken count on \textbf{\framework-MT61}, averaged across all models. Marker shapes indicate writing script. Figures~\ref{fig:subtokens_f200} \& \ref{fig:subtokens_bq275} in the Appendix show \textbf{\framework-F200/BQ275} results.}
%  \label{fig:performance_to_subtokens}
%\end{minipage}
%\hfill
%\begin{minipage}{0.49\textwidth}
%  \centering
%  \includegraphics[width=0.98\linewidth, trim=0 50 0 48, clip]{romansh_heatmap_wide_001.pdf}
%  \caption{\framework MT61 Macro-TACC heatmap per model across Romansh variants. Speaker count according to~\citet{Gross2004-bi} in the year 2000.}
%  \label{fig:single_lang_variants_heatmap}
%\end{minipage}
%\end{figure*}

\begin{figure*}
  \centering
  \includegraphics[width=0.98\linewidth, trim=0 50 0 48, clip]{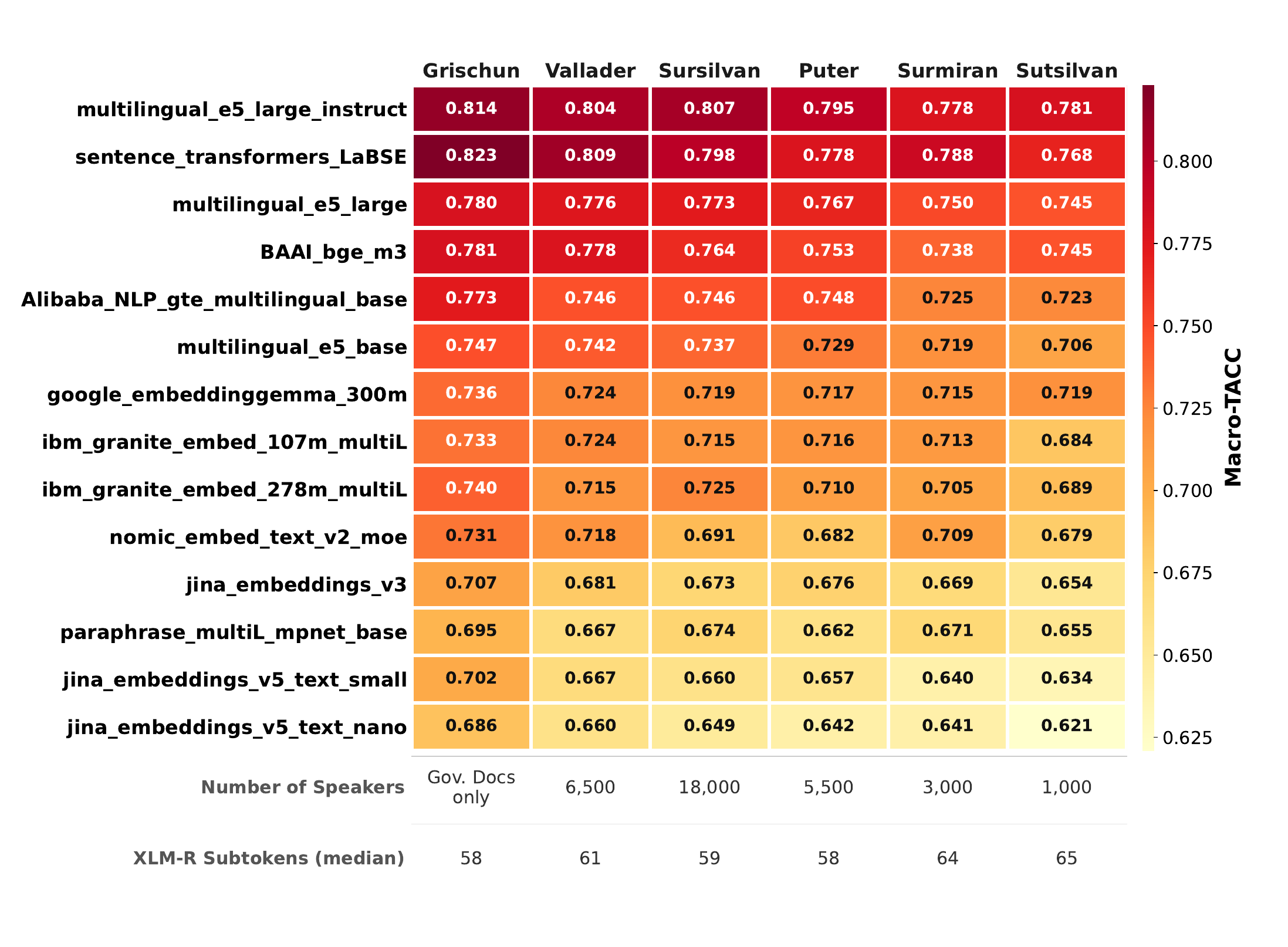}
  \caption{\framework MT61 Macro-TACC for Romansh variants. Speaker count according to~\citet{Gross2004-bi}; year 2000.}
  \label{fig:single_lang_variants_heatmap}
\end{figure*}

\subsection{Case Study: Romansh}

\paragraph{Research Question.} We finally examine an extreme low-resource case: how do multilingual embedding models perform across the six written Romansh varieties?
Romansh is a Rhaeto-Romance language and the fourth national language of Switzerland, with roughly 60,000 speakers in total. It comprises five traditional written varieties, or “idioms”: \textit{Sursilvan}, \textit{Sutsilvan}, \textit{Surmiran}, \textit{Puter}, and \textit{Vallader}, which differ substantially in orthography and phonology \citep{Gross2004-bi}. \textit{Rumantsch Grischun} \citep{alma990033706030205600} is a standardized written form designed for inter-variety communication and official use \citep{Gross2004-bi,Ayres-Bennett2018-ol,vamvas-etal-2025-expanding}. Evaluating these varieties allows us to test model behavior in a fine-grained and low-resource setting.

\paragraph{Setup.} We calculate, for all embedding models and all six Romansh varieties, the average Macro-TACC over all four minimal pair types.

\paragraph{Findings.} Figure~\ref{fig:single_lang_variants_heatmap} shows that performance is highest for Rumantsch Grischun and then generally declines across the spoken varieties in roughly the order of speaker population. This pattern is also accompanied by increasing subtoken fragmentation. The result is consistent with earlier analyses suggesting that written prevalence and tokenizer coverage are associated with better performance. Relative to each model’s overall average across languages, performance on Romansh is typically 10–20\% lower. There are also some notable ranking shifts: for example, mGTE rises to fifth place, and mE5-instruct surpasses LaBSE on \textbf{\framework-MT61} for this subset, suggesting that model rankings can change substantially in very low-resource settings.
Overall, this suggests that \framework could help inform the selection of an embedding model for a low-resource language or dialect of interest, since the performance of embedding models on different languages is not directly indicative of their performance on a specific low-resource language.

\subsection{Decoder-Based Embedding Models}

\paragraph{Research Question.} %Recent embedding models increasingly use decoder-based LLM backbones that are larger than the encoders examined so far. 
We ask whether recent decoder-based LLM embeddings can improve cross-lingual discrimination on \framework.

\paragraph{Setup.} We evaluate the Qwen3-Embedding family (0.6B, 4B, 8B) \citep{zhang2025qwen3embeddingadvancingtext} under the same TACC protocol, comparing against some of the sub-600M encoder models (Table~\ref{tab:qwen_main}).

\paragraph{Findings.} Scale helps within the family: Qwen3-Embedding rises from 0.691 (0.6B) to 0.752 (4B) to 0.801 (8B). But it does not close the gap to encoders. The largest decoder (8B) ranks sixth overall, below LaBSE (0.879) despite roughly an order of magnitude more parameters, and is matched or beaten by smaller encoder models (Table~\ref{tab:qwen_main}). Parameter count is thus not the primary driver here: the tested decoder-based models do not appear to capture the fine-grained cross-lingual contrasts in \framework as reliably as smaller encoders, which remain more effective despite their size. The pattern holds across all three datasets (per-augmentation breakdowns are reported in Appendix Table~\ref{tab:qwen_appendix}).

\section{Conclusions}
We present \framework, a dynamic cross-lingual evaluation framework that generates fine-grained semantic stress tests for embedding models in any language with English parallel data.
Across 275+ languages—many previously lacking embedding evaluation resources—we find (1) persistent gaps in cross-lingual semantic representation; (2) that explicit semantic reversals such as negation are easier to detect than structural shifts like role swaps or abstraction-level changes; (3) and that paragraph-level texts are more challenging than single sentences. Further, (4) performance is strongly tied to language prevalence in training data and tokenizer coverage, a pattern that holds down to the dialect level, as shown in our Romansh case study. Model rankings also shift across languages, meaning aggregate scores can obscure important per-language differences. Beyond these findings, \framework provides dynamic evaluation for numerous languages, including many low-resource varieties not previously covered by embedding benchmarks. %We release the framework and data to support evaluation of current and future models.

\section*{Acknowledgments}
This research is conducted under the project \textit{Impresso -- Media Monitoring of the Past II. Beyond Borders: Connecting Historical Newspapers and Radio}. Impresso is a research project funded by the Swiss National Science Foundation (SNSF 213585) and the Luxembourg National Research Fund (17498891). MW acknowledges funding by the Swiss National Science Foundation (project InvestigaDiff; no.\ 10000503).

\section*{Limitations}

Our framework applies semantic perturbations on the English side of parallel data, enabling precise and controllable edits through AMR. As a result, it evaluates cross-lingual alignment to semantic contrasts expressed in English, rather than performing edits directly within the target language. While this introduces an English-centric perspective, it provides a consistent and scalable evaluation setup; extending controlled perturbations to additional languages is a promising direction for future work.

%The framework assumes that target-language sentences are faithful translations of the English source. Although we rely on high-quality parallel corpora and observe consistent trends across datasets, translation noise—especially in low-resource settings—may affect results. Incorporating translation quality estimation or filtering could further improve robustness.

The dynamically generated benchmark's quality depends on the AMR parser, generation model, and NLI-based filtering. While manual inspection suggests high validity, some noise may remain. In addition, certain perturbations (e.g., role swaps) can produce less natural sentences, which can also be interpreted as testing robustness to atypical or adversarial inputs. Advances in generation and validation could further improve naturalness and control.

Our evaluation uses a pairwise ranking metric (Triplet Accuracy), which provides a clear signal of relative preferences but does not assess absolute similarity calibration. Complementary evaluation metrics could provide a more complete picture.

%Finally, \framework relies on English-centric parallel data and currently covers four perturbation types. While this enables broad multilingual evaluation of core semantic distinctions, future work could expand both language coverage (e.g., beyond English pivots) and the range of semantic transformations. The framework is readily extensible to new models and settings.

\bibliography{custom,bib_from_ss}

\appendix

\section{Appendix}

% Paste this into your LaTeX document (requires booktabs, multirow, adjustbox packages)
% \usepackage{booktabs, multirow, adjustbox, caption}

\begin{table*}[t]
\centering
\footnotesize
\setlength{\arrayrulewidth}{0.3pt}
\resizebox{\textwidth}{!}{%
\begin{tabular}{lll}
\toprule
\textbf{Model Short Name} & \textbf{Hugging Face ID} & \textbf{Instructions / Prompt / Prefix / Adapter} \\
\midrule
sentence\_transformers\_LaBSE       & \texttt{sentence-transformers/LaBSE}                              & ---                                                                \\ \hline
BAAI\_bge\_m3                       & \texttt{BAAI/bge-m3}                                              & ---                                                                \\ \hline
multilingual\_e5\_large\_instruct   & \texttt{intfloat/multilingual-e5-large-instruct}                   & \{Prefix:\} \texttt{Instruct: Represent every}                       \\
                                    &                                                                    & \texttt{detail of this text to avoid}                                \\
                                    &                                                                    & \texttt{matching to hard negatives.\textbackslash nQuery:}                    \\ \hline
Qwen3-Embedding-0.6B   & \texttt{Qwen/Qwen3-Embedding-0.6B}                   & \{Prefix:\} \texttt{Instruct: Represent every}                       \\
                                    &                                                                    & \texttt{detail of this text to avoid}                                \\
                                    &                                                                    & \texttt{matching to hard negatives.\textbackslash nQuery:}                    \\ \hline
Qwen3-Embedding-4B   & \texttt{Qwen/Qwen3-Embedding-4B}                   & \{Prefix:\} \texttt{Instruct: Represent every}                       \\
                                    &                                                                    & \texttt{detail of this text to avoid}                                \\
                                    &                                                                    & \texttt{matching to hard negatives.\textbackslash nQuery:}                    \\ \hline
Qwen3-Embedding-8B   & \texttt{Qwen/Qwen3-Embedding-8B}                   & \{Prefix:\} \texttt{Instruct: Represent every}                       \\
                                    &                                                                    & \texttt{detail of this text to avoid}                                \\
                                    &                                                                    & \texttt{matching to hard negatives.\textbackslash nQuery:}                    \\ \hline
multilingual\_e5\_large             & \texttt{intfloat/multilingual-e5-large}                            & \{Prefix:\} \texttt{query:}                                            \\ \hline
multilingual\_e5\_base              & \texttt{intfloat/multilingual-e5-base}                             & \{Prefix:\} \texttt{query:}                                            \\ \hline
jina\_embeddings\_v3                & \texttt{jinaai/jina-embeddings-v3}                                 & \{Adapter:\} \texttt{text-matching}                                    \\ \hline
google\_embeddinggemma\_300m        & \texttt{google/embeddinggemma-300m}                                & Prefix: \texttt{task: sentence}                                    \\
                                    &                                                                    & \texttt{similarity | query:}                                       \\ \hline
nomic\_embed\_text\_v2\_moe         & \texttt{nomic-ai/nomic-embed-text-v2-moe}                          & \{Prefix:\} \texttt{search\_query:}                                    \\ \hline
jina\_embeddings\_v5\_text\_small   & \texttt{jinaai/jina-embeddings-v5-text-small}                      & \{Adapter:\} \texttt{text-matching}                                    \\ \hline
paraphrase\_multiL\_mpnet\_base     & \texttt{sentence-transformers/paraphrase-}                         & ---                                                                \\
                                    & \texttt{multilingual-mpnet-base-v2}                                &                                                                    \\ \hline
jina\_embeddings\_v5\_text\_nano    & \texttt{jinaai/jina-embeddings-v5-text-nano}                       & \{Adapter:\} \texttt{text-matching}                                    \\ \hline
gte-multilingual-base     & \texttt{Alibaba-NLP/gte-multilingual-base}                         & ---                                                                \\ \hline
ibm\_granite\_embed\_107m\_multiL   & \texttt{ibm-granite/granite-embedding-107m-multilingual}           & --- \\ \hline
ibm\_granite\_embed\_278m\_multiL   & \texttt{ibm-granite/granite-embedding-278m-multilingual}           & ---                                                                \\ \hline                    
\bottomrule
\end{tabular}%
}
\caption{Embedding model configurations used in the evaluation.}
\label{tab:model-configs}
\end{table*}
\label{sec:appendix}

\begin{table*}[t]
\centering
\footnotesize
\setlength{\tabcolsep}{2.3pt}
\resizebox{\textwidth}{!}{%
\begin{tabular}{l rrrrr | rrrrr | rrrrr}
\toprule
Model & \multicolumn{5}{c|}{\textbf{\framework-F200}} & \multicolumn{5}{c|}{\textbf{\framework-MT61}} & \multicolumn{5}{c}{\textbf{\framework-BQ275}} \\
& PN & RS & AR & HS & Avg & PN & RS & AR & HS & Avg & PN & RS & AR & HS & Avg \\
\midrule
multilingual\_e5\_large\_instruct & 0.945 & 0.882 & 0.904 & 0.889 & 0.905 & 0.929 & 0.872 & 0.899 & 0.895 & 0.899 & 0.892 & 0.821 & 0.835 & 0.821 & 0.842 \\
sentence\_transformers\_LaBSE & 0.896 & 0.903 & 0.884 & 0.903 & 0.896 & 0.920 & 0.913 & 0.910 & 0.925 & 0.917 & 0.807 & 0.828 & 0.810 & 0.847 & 0.823 \\
multilingual\_e5\_large & 0.920 & 0.873 & 0.889 & 0.878 & 0.890 & 0.895 & 0.860 & 0.891 & 0.885 & 0.883 & 0.821 & 0.795 & 0.761 & 0.783 & 0.790 \\
BAAI\_bge\_m3 & 0.917 & 0.867 & 0.881 & 0.884 & 0.887 & 0.888 & 0.866 & 0.890 & 0.896 & 0.885 & 0.828 & 0.767 & 0.771 & 0.764 & 0.782 \\
multilingual\_e5\_base & 0.912 & 0.860 & 0.880 & 0.852 & 0.876 & 0.835 & 0.824 & 0.852 & 0.858 & 0.842 & 0.801 & 0.769 & 0.747 & 0.756 & 0.768 \\
\textbf{Qwen3\_Embedding\_8B} & 0.880 & 0.821 & 0.836 & 0.833 & 0.842 & 0.872 & 0.810 & 0.847 & 0.826 & 0.839 & 0.778 & 0.707 & 0.710 & 0.697 & 0.723 \\
Alibaba\_NLP\_gte\_multilingual\_base & 0.849 & 0.806 & 0.825 & 0.793 & 0.818 & 0.828 & 0.785 & 0.827 & 0.837 & 0.819 & 0.808 & 0.743 & 0.733 & 0.755 & 0.760 \\
google\_embeddinggemma\_300m & 0.871 & 0.765 & 0.798 & 0.775 & 0.802 & 0.881 & 0.826 & 0.852 & 0.838 & 0.849 & 0.807 & 0.675 & 0.699 & 0.661 & 0.711 \\
nomic\_embed\_text\_v2\_moe & 0.788 & 0.797 & 0.807 & 0.837 & 0.807 & 0.796 & 0.818 & 0.824 & 0.850 & 0.822 & 0.725 & 0.737 & 0.710 & 0.754 & 0.732 \\
jina\_embeddings\_v3 & 0.832 & 0.729 & 0.756 & 0.732 & 0.762 & 0.869 & 0.799 & 0.835 & 0.808 & 0.828 & 0.791 & 0.651 & 0.685 & 0.656 & 0.696 \\
\textbf{Qwen3\_Embedding\_4B} & 0.825 & 0.754 & 0.766 & 0.762 & 0.777 & 0.828 & 0.773 & 0.795 & 0.801 & 0.799 & 0.755 & 0.645 & 0.667 & 0.656 & 0.680 \\
paraphrase\_multiL\_mpnet\_base & 0.811 & 0.700 & 0.759 & 0.728 & 0.750 & 0.835 & 0.776 & 0.826 & 0.810 & 0.812 & 0.756 & 0.647 & 0.675 & 0.664 & 0.686 \\
jina\_embeddings\_v5\_text\_small & 0.797 & 0.744 & 0.757 & 0.751 & 0.762 & 0.824 & 0.786 & 0.816 & 0.816 & 0.810 & 0.706 & 0.648 & 0.619 & 0.617 & 0.648 \\
ibm\_granite\_embed\_278m\_multiL & 0.744 & 0.735 & 0.718 & 0.726 & 0.731 & 0.781 & 0.765 & 0.782 & 0.797 & 0.781 & 0.708 & 0.685 & 0.654 & 0.695 & 0.685 \\
jina\_embeddings\_v5\_text\_nano & 0.797 & 0.724 & 0.736 & 0.708 & 0.741 & 0.811 & 0.777 & 0.792 & 0.786 & 0.791 & 0.709 & 0.661 & 0.613 & 0.602 & 0.646 \\
ibm\_granite\_embed\_107m\_multiL & 0.732 & 0.726 & 0.697 & 0.709 & 0.716 & 0.765 & 0.743 & 0.753 & 0.775 & 0.759 & 0.699 & 0.672 & 0.631 & 0.668 & 0.667 \\
\textbf{Qwen3\_Embedding\_0.6B} & 0.715 & 0.682 & 0.682 & 0.704 & 0.696 & 0.737 & 0.725 & 0.735 & 0.742 & 0.735 & 0.666 & 0.631 & 0.630 & 0.648 & 0.644 \\
\bottomrule
\end{tabular}%
}
\caption{Per-category Macro-TACC results across all three \framework datasets.}
\label{tab:qwen_appendix}
\end{table*}

\begin{figure*}[t]
\centering
\includegraphics[width=0.98\linewidth, trim={0 0.5cm 0 0}, clip]{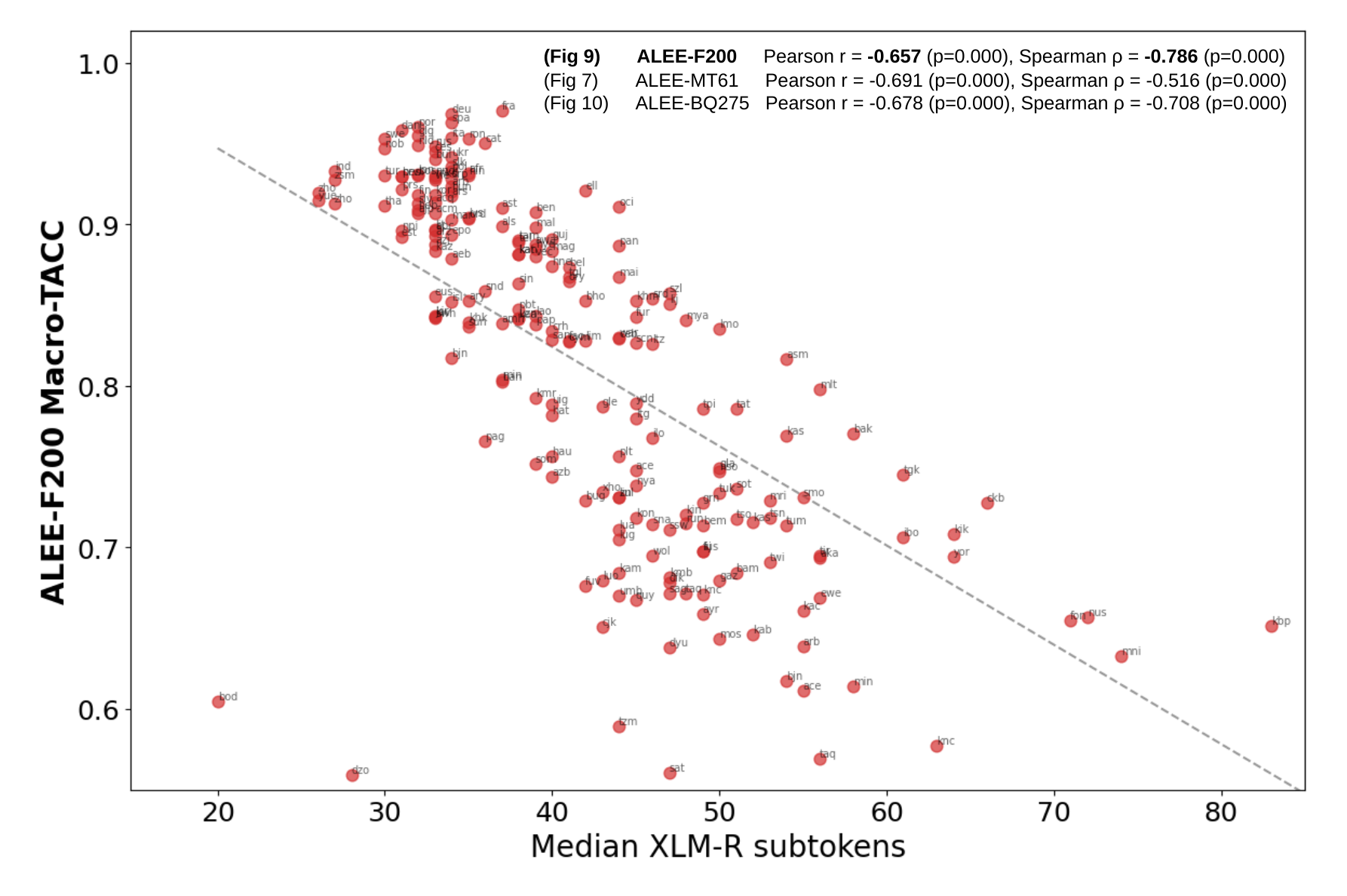}
\caption{Per-language Macro-TACC vs.\ median XLM-R subtoken count on \textbf{\framework-F200}, averaged across all models. Marker shapes indicate writing script.}
\label{fig:subtokens_f200}
\end{figure*}

\begin{figure*}[t]
\centering
\includegraphics[width=0.98\linewidth, trim={0 0.5cm 0 0}, clip]{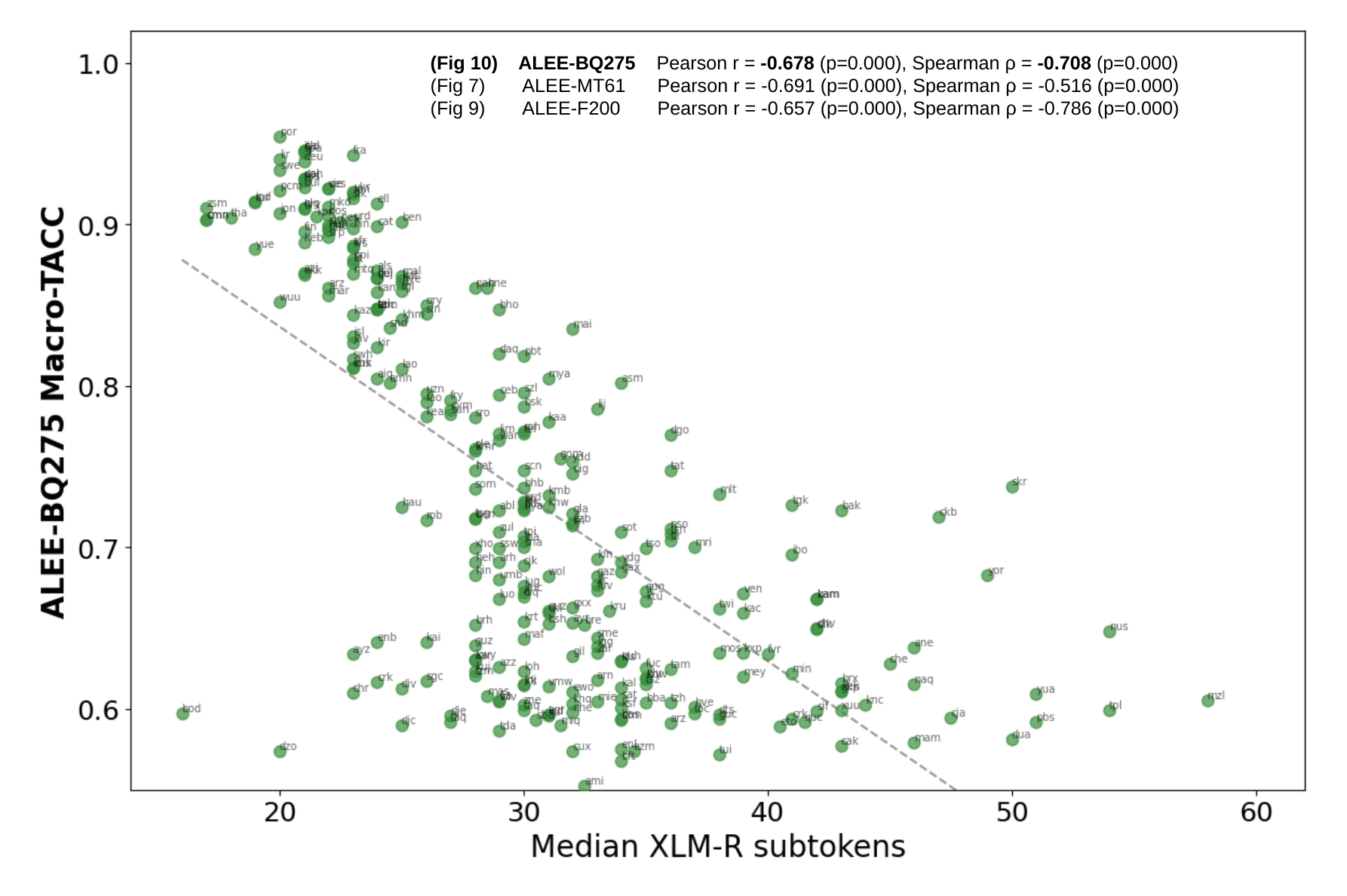}
\caption{Per-language Macro-TACC vs.\ median XLM-R subtoken count on \textbf{\framework-BQ275}, averaged across all models. Marker shapes indicate writing script.}
\label{fig:subtokens_bq275}
\end{figure*}

\begin{figure*}[t]
\centering
\includegraphics[width=0.99\linewidth, trim=26 0 26 0, clip]{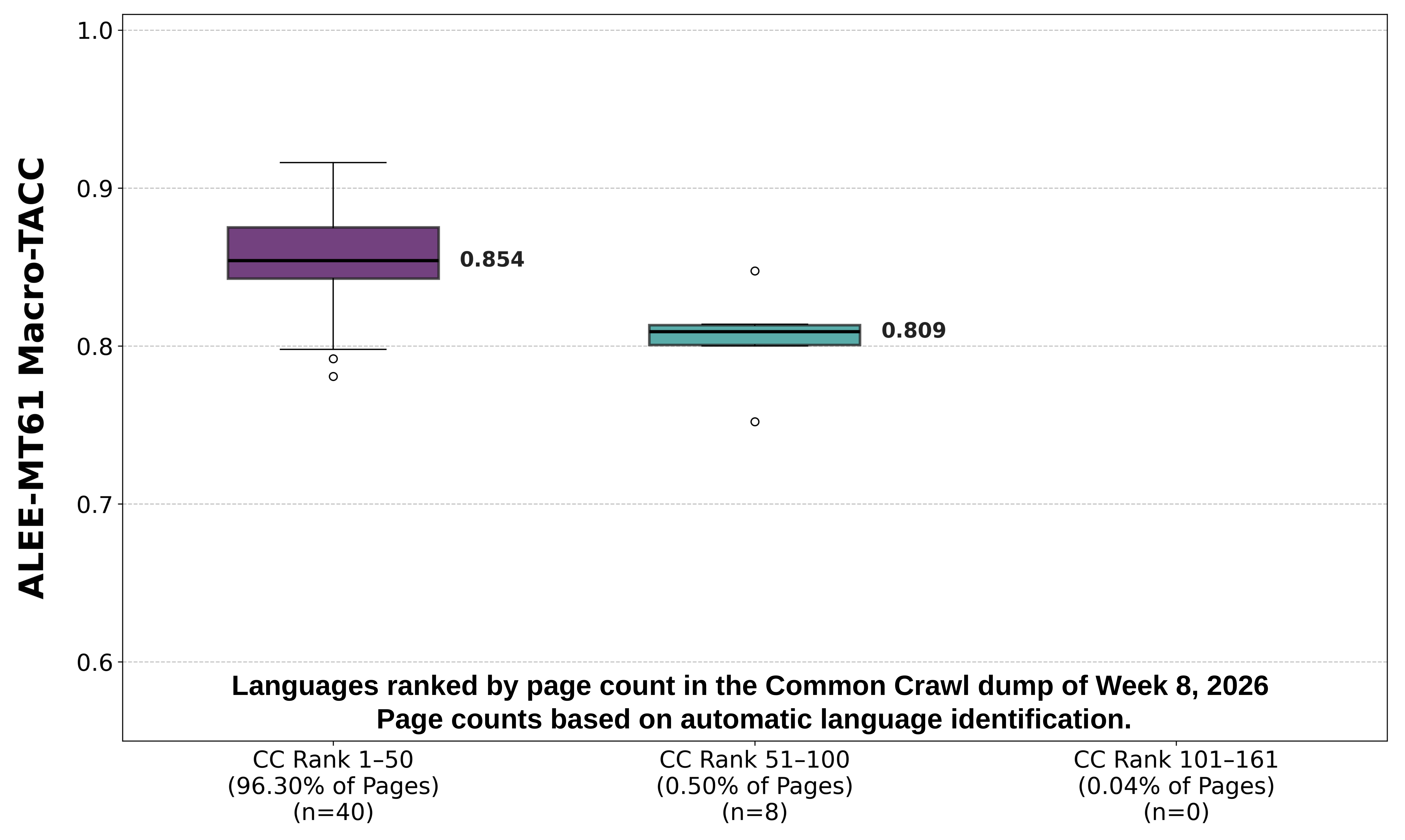}
\caption{\framework-MT61 Macro-TACC distribution by Common Crawl language prevalence. Analogous to Figure~\ref{fig:resource_group_boxplot}.}
\label{fig:resource_group_boxplot_mt61}
\end{figure*}

\begin{figure*}[t]
\centering
\includegraphics[width=0.99\linewidth, trim=26 0 26 0, clip]{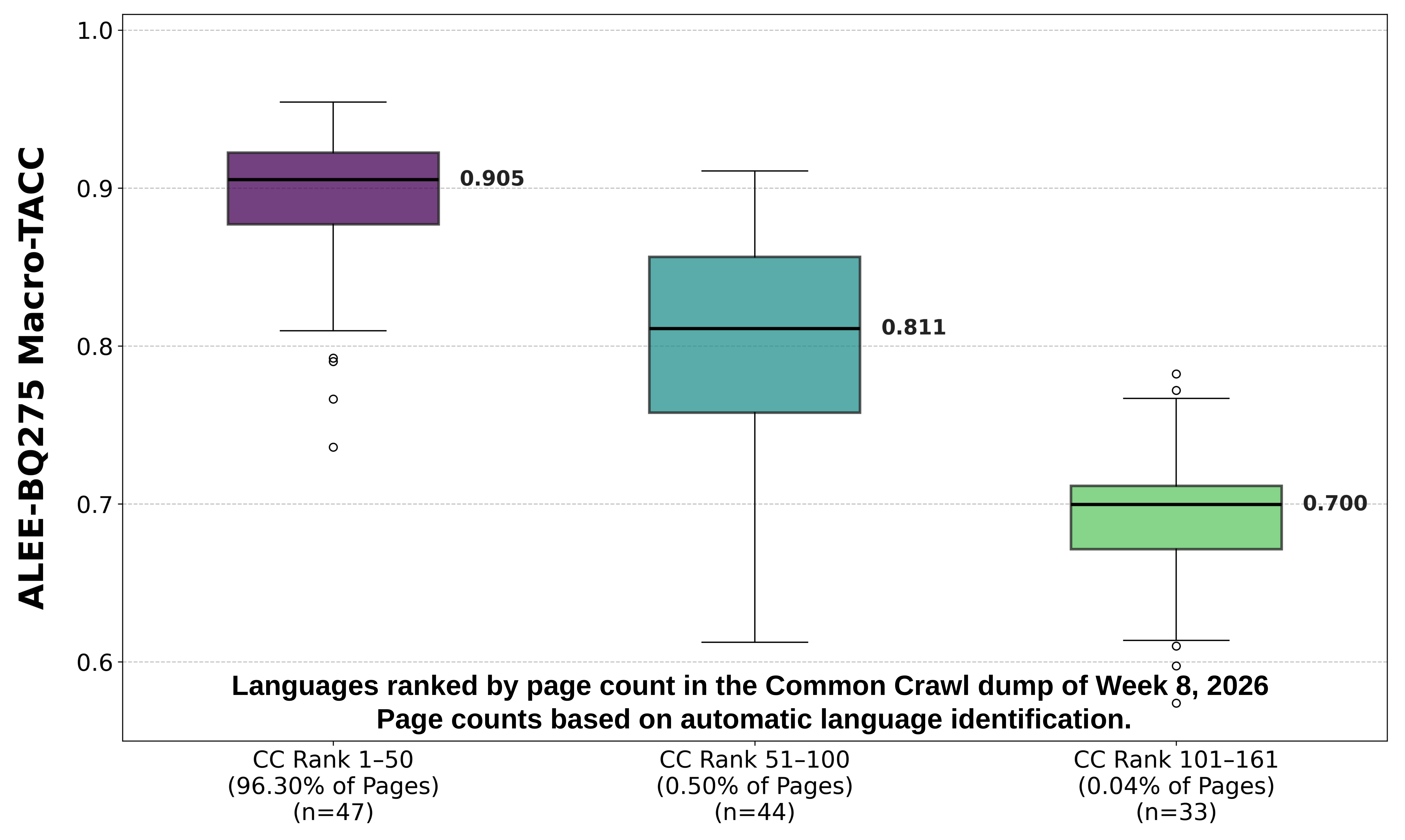}
\caption{\framework-BQ275 Macro-TACC distribution by Common Crawl language prevalence. Analogous to Figure~\ref{fig:resource_group_boxplot}.}
\label{fig:resource_group_boxplot_bq275}
\end{figure*}

% Requires booktabs and array packages
% \usepackage{booktabs, array}

\clearpage

\section{Appendix: Per Language Results}
\label{app_per_language}
\input{appendix_tables_output}

\end{document}